\documentclass[10pt,twocolumn,letterpaper]{article}

\usepackage{wacv}
\usepackage{times}
\usepackage{epsfig}
\usepackage{graphicx}
\usepackage{subfigure}
\usepackage{amsmath}
\usepackage{amssymb}
\usepackage{bbm}
\usepackage{bm}
\usepackage{color}
\usepackage{url}
\usepackage[colorlinks=true]{hyperref}

\wacvfinalcopy 

\def\wacvPaperID{232} 

\ifwacvfinal\fi
\begin{document}

\title{Deep Learning Logo Detection with Data Expansion by Synthesising Context}

\author{Hang Su \hspace{2cm} Xiatian Zhu \hspace{2cm} Shaogang Gong \\
Shchool of EECS, Queen Mary University of London, United Kingdom \\
{\tt\small hang.su@qmul.ac.uk, xiatian.zhu@qmul.ac.uk, s.gong@qmul.ac.uk}
}

\maketitle
\ifwacvfinal\thispagestyle{empty}\fi

\begin{abstract}
Logo detection in unconstrained images is challenging,
particularly when only very sparse labelled training images are
accessible due to high labelling costs.
In this work, we describe a model training image synthesising method 
capable of improving significantly logo detection performance when only 
a handful of (e.g., 10) labelled training images captured in realistic context
are available, avoiding extensive manual labelling costs.
Specifically, we design a novel algorithm for generating 
Synthetic Context Logo (SCL) training images to increase model
robustness against unknown background clutters, resulting in superior
logo detection performance.
For benchmarking model performance, we introduce a new logo detection
dataset TopLogo-10 collected from top 10 most popular
clothing/wearable brandname logos captured in
rich visual context. 
Extensive comparisons show the advantages of our proposed SCL model
over the state-of-the-art alternatives for logo detection using two
real-world logo benchmark datasets: FlickrLogo-32 and our new TopLogo-10\footnote{
The TopLogo-10 dataset is available at: \url{http://www.eecs.qmul.ac.uk/~hs308/qmul_toplogo10.html}}.


\end{abstract}

\section{Introduction}


Logo detection is a challenging task for computer vision,
with a wide range of applications in many domains, 
such as brand logo recognition for commercial research, 
brand trend research on Internet social community, 
vehicle logo recognition for intelligent transportation
\cite{romberg2011scalable,revaud2012correlation,romberg2013bundle,boia2014local,li2014logo,pan2013vehicle}. 
%
%
For generic object detection, deep learning 
has been a great success 
\cite{ren2015faster,redmon2015you,shrivastava2016training}.
Building a deep object detection model typically requires a large
number of labelled training data collected from extensive
manual labelling \cite{krizhevsky2012imagenet,simonyan2014very}.
However, this is not necessarily available in many cases such as logo detection where
the publicly available datasets are very small (Table
\ref{tab:dataset}). Small training data size is inherently inadequate
for learning deep models with millions of parameters.
Increasing manual annotation is extremely costly
\cite{hoi2015logo} and unaffordable in most cases, not only in
monetary but more critically in timescale terms.
 
\begin{figure}[]
	\centering
	\includegraphics[width=0.495\linewidth]{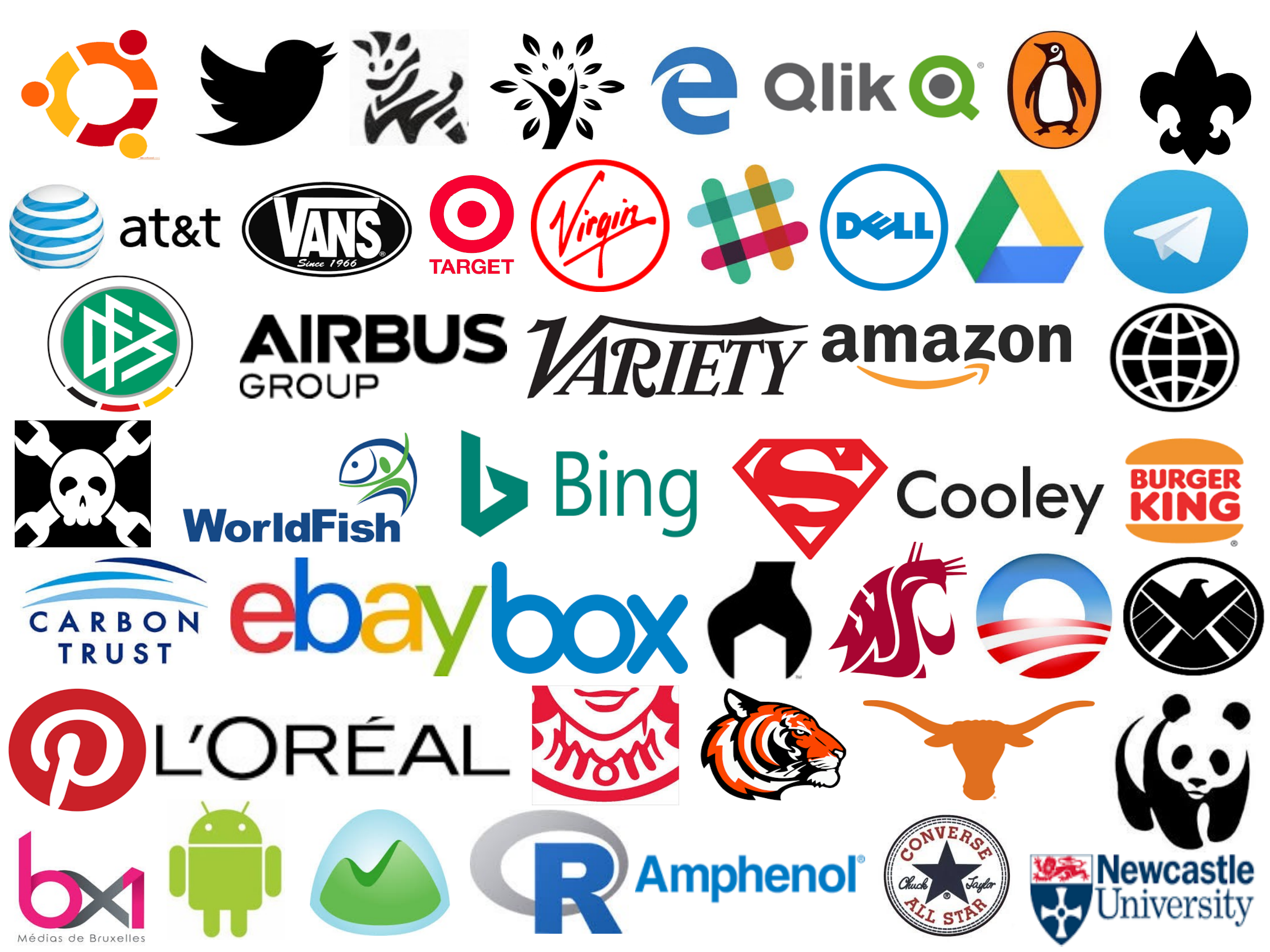}
	\includegraphics[width=0.495\linewidth]{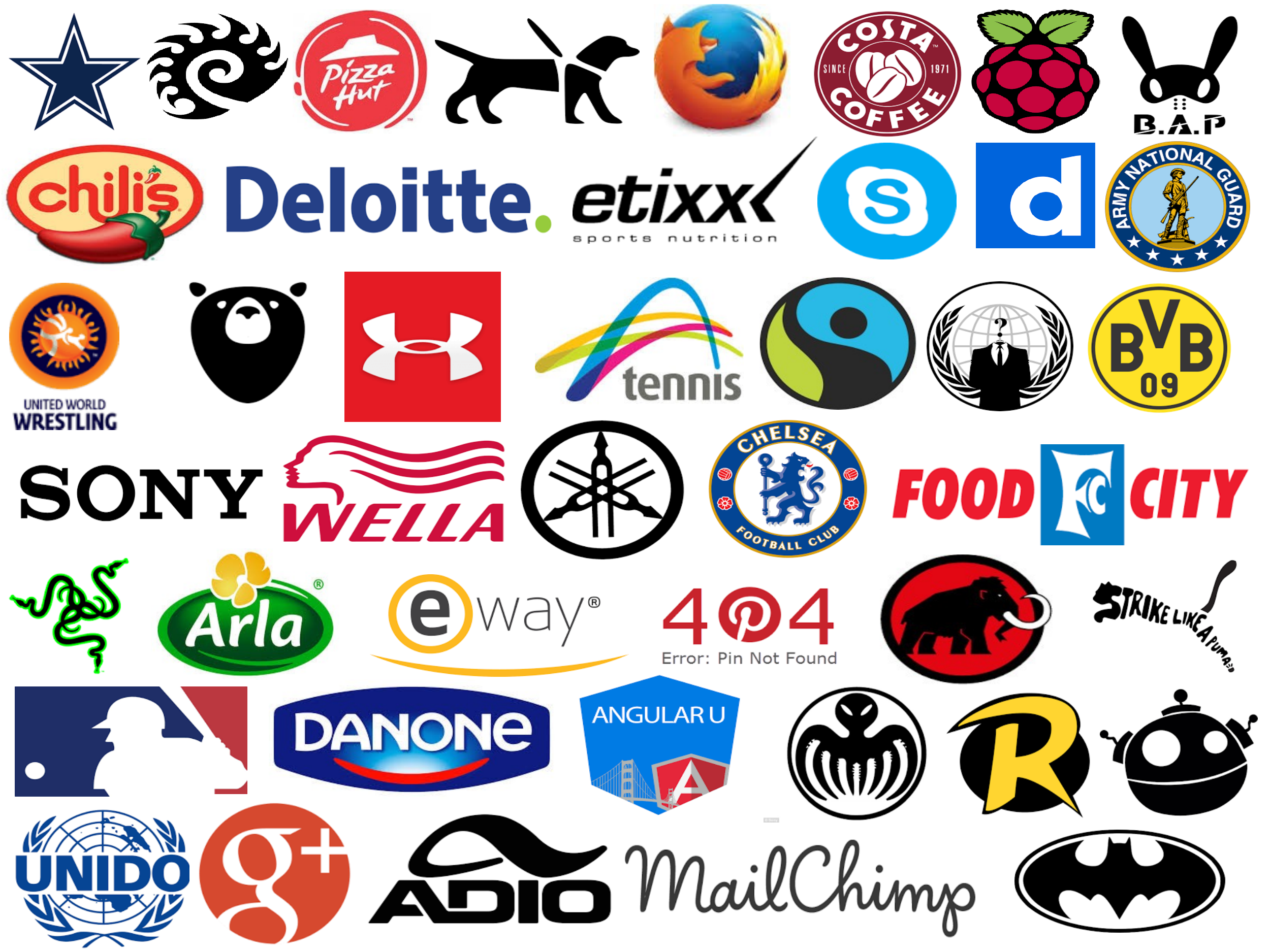}
	\includegraphics[width=0.495\linewidth]{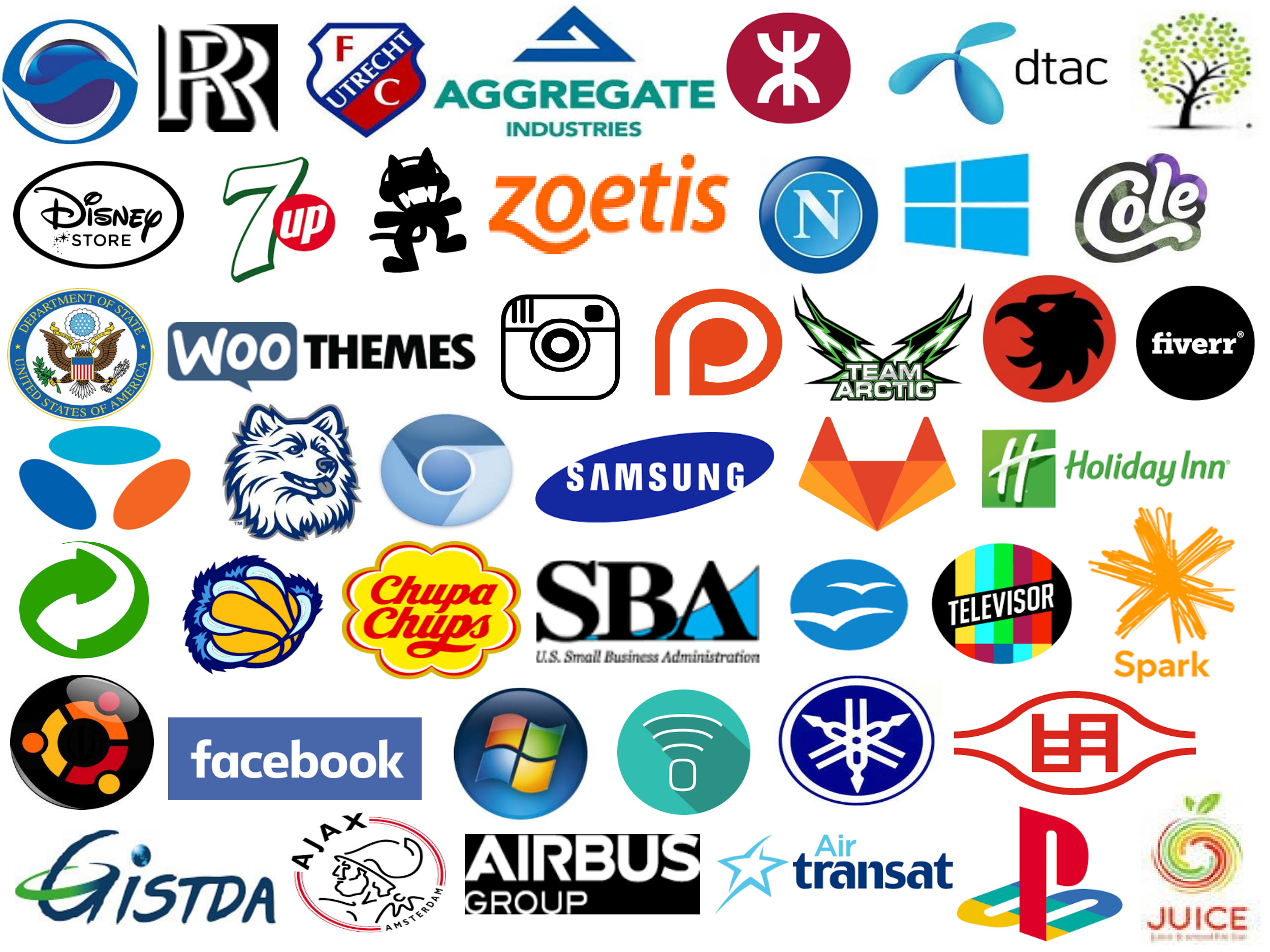}
	\includegraphics[width=0.495\linewidth]{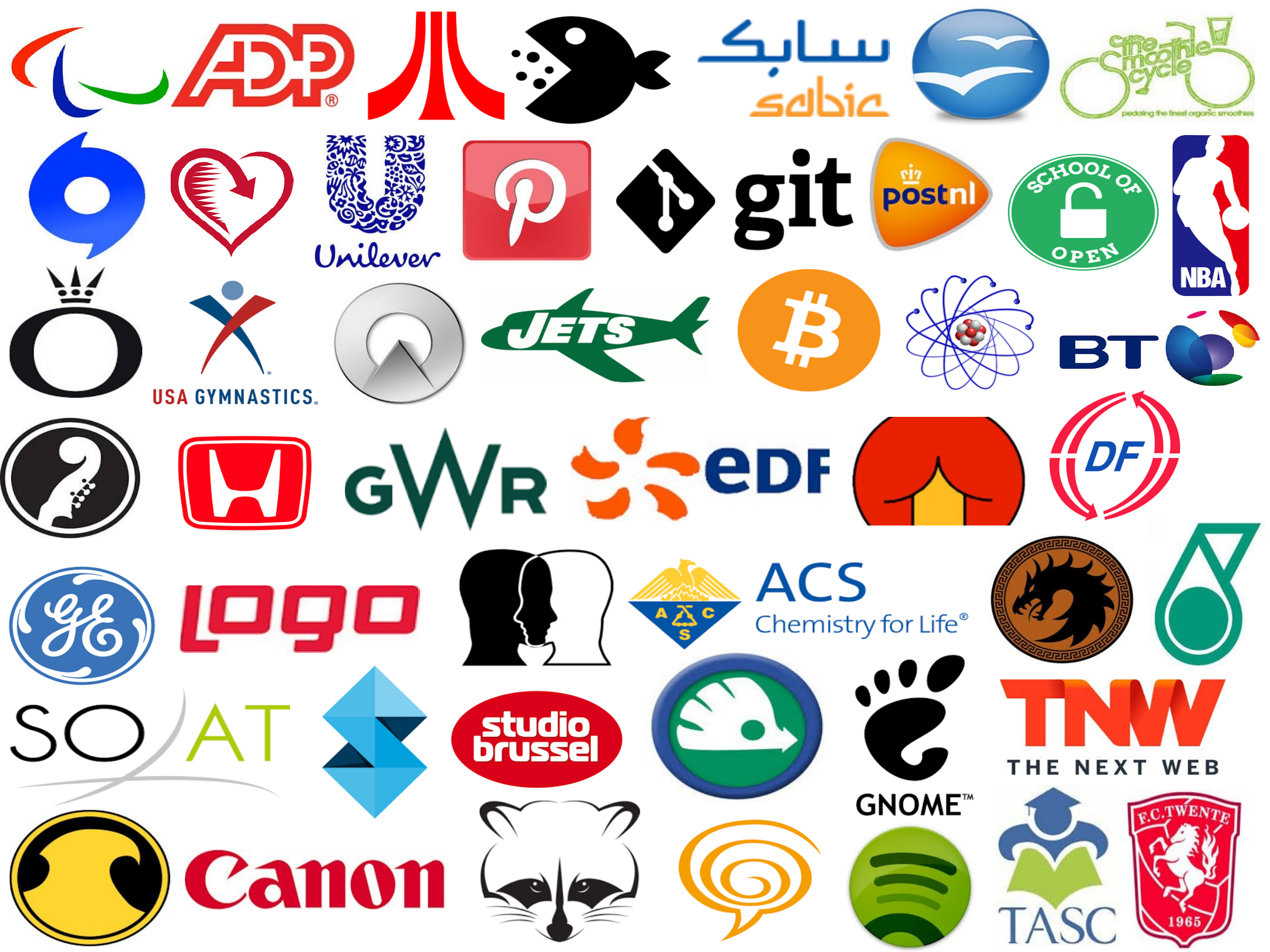}
	\caption{Examples of logo exemplar images.}
	\label{fig:logo_examples}
\end{figure}


In the current literature, most existing studies on logo detection 
are limited to small scales, in both the number of logo images and
logo classes \cite{iandola2015deeplogo,li2014logo},
%
largely due to the high costs in constructing large scale logo datasets.
It is non-trivial to collect automatically large scale logo training
data that covers a large number of different logo classes. 
While web data mining may be a potential solution as shown in other recognition problems \cite{Liang_2015_ICCV,chen2015webly,chen2013neil,shrivastava2016training},
it is difficult to acquire accurate logo annotations since no
bounding box annotation is available from typical web images
and their meta-data.
%

\begin{table}  %
	\footnotesize
	\centering
	\caption{
		Existing logo detection datasets.
		PA: Public Availability.
		Apart from labelled logo images, 
		BelgaLogos provides 8,049 images with no bounding boxes
		and FlickrLogos-32 has 6,000 non-logo images.	
	}
	\vskip 0.1cm
	\label{tab:dataset}
	\begin{tabular}{c||c|c|c|c}
		\hline
		Dataset & Logo \# & Object \# & Image \# &  PA \\
		\hline \hline
		BelgaLogos \cite{joly2009logo}  
		& 37  & 2,695 & 1,951 & Yes \\
		\hline
		FlickrLogos-27 \cite{kalantidis2011scalable}
		& 27 & 4,536 & 810 & Yes \\
		\hline
		FlickrLogos-32 \cite{romberg2011scalable}
		& 32 & 3404 & 2,240 & Yes \\
		\hline
		LOGO-NET \cite{hoi2015logo}
		& 160 & 130,608 & 73,414  & No \\
		\hline
	\end{tabular}
\end{table}



In this work, we present a novel synthetic training data generation
algorithm for improving the learning of a deep logo detector with only 
sparsely labelled training images. 
This approach enlarges significantly the variations of both logo and
its context in the training data without increasing manual labelling
effort, so that a deep detector can be optimised 
to recognise the target logos against diverse and complex background
clutters not captured by the original sparse training data.
%
%
The {\em contributions} of this work are:
(1) We formulate a novel Synthetic Context Logo (SCL) training data generation method 
for learning a logo detector given sparsely labelled images. 
Unlike typical deep learning models,
we do not assume the availability of large quantities of labelled training images 
but only a handful. Our model is designed specifically to augment the training data
by enriching and expanding both logos and their context variations.
To our knowledge, it is the first attempt of exploiting 
large scale synthetic training {\em data expansion in context} for
deep learning a logo detection model.
(2) We introduce a large scale automatically synthesised logo dataset, in rich
context with labelled logo bounding boxes, consisting of $463$
different logos (Figure \ref{fig:logo_examples}) which is much larger
in class number than any existing logo benchmark datasets in the public domain.
(3) We further introduce a new logo dataset TopLogo-10, manually
collected and labelled from in-the-wild logo images. This TopLogo-10
dataset consists of 10 logo classes of high logo popularity (with high
frequency in real-life \cite{top10best,2015brandzluxury,top10fashion,50streetwear}
in rich context, thus more challenging.
We evaluated extensively the proposed SCL method for
deep learning a logo detector against the state of the art
alternatives using two logo benchmark datasets, and
provided in-depth analysis and discussion on model generalisation.


\section{Related Works}
\label{related works}


\noindent {\bf Logo Detection.} Most existing approaches to logo detection rely on 
hand-crafted features, e.g., HOG, SIFT, colour histogram, edge
\cite{kalantidis2011scalable,romberg2011scalable,revaud2012correlation,romberg2013bundle,boia2014local,li2014logo,pan2013vehicle}. 
They are limited in obtaining more expressive representation and
model robustness for recognising a large number of different logos. 
One reason is due to the unavailability of sufficiently large datasets 
required for exploring deep learning a more discriminative representation.
For example, among all publicly available logo datasets,  
%
%
the most common FlickrLogos-32 dataset \cite{romberg2011scalable}
contains $32$ logo classes each with only $70$ images
and in total $5644$ logo objects,
whilst BelgaLogos \cite{joly2009logo} has $2695$ logo images from
$37$ logo classes with bounding box (bbox) location labelled
(Table \ref{tab:dataset}).
%
%
%
%
%
%
%
%
%


Nevertheless, a few deep learning based
logo detection models have been reported recently.
%
Iandola et al. \cite{iandola2015deeplogo} 
applied the Fast R-CNN model \cite{girshick2015fast} for logo detection,
which inevitably suffers from the training data scarcity challenge.
To facilitate deep learning logo detection,
Hoi et al. \cite{hoi2015logo} built a large scale dataset
called LOGO-Net by exhaustively collecting images from 
online retailer websites and then manually labelling them.
This requires a huge amount of construction effort and moreover, LOGO-Net is inaccessible publicly.
In contrast to all existing attempts above, we explore the potentials 
for learning a deep logo detection model
by synthesising a large scale sized training data to address the
sparse data annotation problem without additional human labelling cost.
Compared to \cite{hoi2015logo}, our method is much more cost-effective and
scalable for logo variations in diverse visual context, e.g., accurate
logo annotation against diverse visual scene context can be rapidly
generated without any manual labelling, and potentially also for
generalising to a large number of new logo classes with minimal labelling.


\vspace{0.2cm}
\noindent {\bf Synthesising Data Expansion.} Generating synthetic
training data allows for expanding plausibly ground-truth annotations 
without the need for exhaustive manual labelling.
This strategy has been shown to be effective for training large CNN models 
particularly when no sufficient training data are available,
%
e.g., Dosovitskiy et al. \cite{dosovitskiy2015flownet} 
used synthetic floating chair images for training optical flow networks;
Gupta et al. \cite{gupta2016synthetic}
and Jaderberg et al. \cite{jaderberg2016reading}
generated scene-text images for learning text recognition models;
%
%
Yildirim et al. \cite{yildirim2015efficient} 
exploited deep CNN features optimised on synthetic faces 
to regress face pose parameters.
Eggert et al. \cite{eggert2015benefit} 
applied synthetic data to train SVM models for 
company logo detection,
which shares the spirit of the proposed method
but with an essential difference in
that we explore synthetic training data {\em in 
diverse context variations} 
in the absence of large scale realistic logo dataset in context.



%

\section{Synthetic Logos in Context}
\label{Dataset}

Typically, a large training dataset is required 
for learning an effective deep network \cite{krizhevsky2012imagenet}.
This however is very expensive to collect manually,
particularly when manual annotation of locations and bounding boxes of varying-sized objects
are needed, e.g., logos in natural images from the wild street views
\cite{hoi2015logo}.
Given the small size labelled training data in existing logo datasets (Table \ref{tab:dataset}),
it is challenging to learn effectively the huge number of parameters in deep models.
To solve this problem, we synthesis additional training data. 
We consider that logo variations are largely due to change of visual
context and/or background plus geometric and illumination
transforms. We then develop a 
{\em Synthetic Context Logo} (SCL) image generation method for
deep learning a logo detection model. By doing so, our model is
capable of automatically generating infinite numbers of synthetic
labelled logo images in realistic context with little supervision,
making deep learning feasible. 
%

In model learning, we exploit a sequential learning strategy by first
deploying a large number of synthesised images to pre-train a deep
model, 
followed by fine-tuning the deep model with the sparse manually annotated images.
This sequential model training strategy echoes the principle of Curriculum
Learning (CL), designed for reducing the difficulties in learning
tasks by easy-to-hard staged learning \cite{bengio2009curriculum}. In
learning a logo model, we consider learning from a small number of real images against
cluttered background captured in the wild is a more challenging learning task than learning
from a large number of synthetic images. We evaluated in our
experiments the effectiveness of this staged learning strategy against
model training based on a fusion of synthetic and manually annotated images (Section \ref{sec:eval}).

\subsection{Logo Exemplar and Context Images}
\label{sec:logo_example}

\begin{figure}[]
	\centering
	\includegraphics[width=1\linewidth]{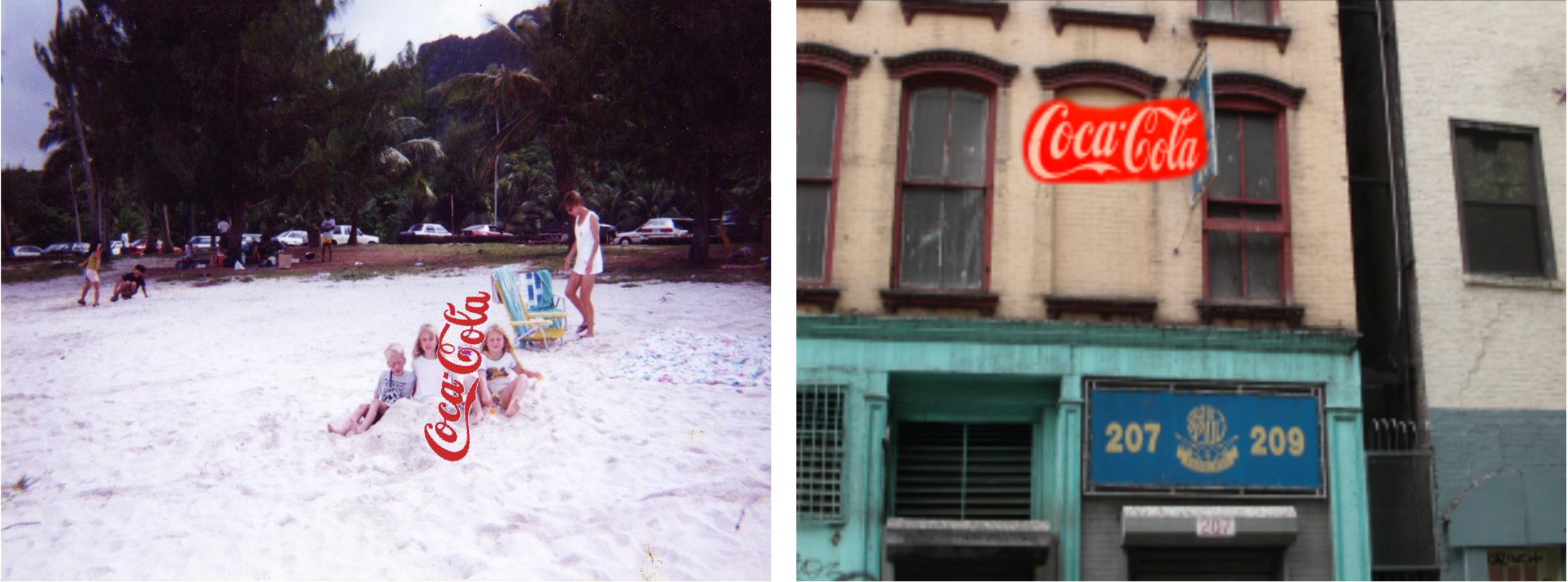}
	\caption{
		Comparing the visual effect of our logo exemplar with transparent background ({\bf left})
		and that of \cite{eggert2015benefit} with non-transparent background ({\bf right})
		in the synthetic logo images.
		Evidently our logo exemplar allows more diverse
                context than \cite{eggert2015benefit}, which imposes a
                surrounding appearance in the synthetic image. More natural appearance is provided by our synthetic image.
		}
	\label{synth background compare}
\end{figure}

To synthesise images for a given logo class, 
we need an exemplar image for each logo class.
This is obtained from Google Image Search with the corresponding logo
name as the query keyword.
To minimise any context bias in the synthetic images,
we utilised those exemplars with pure logo against a homogeneous
transparent background (Figure \ref{fig:logo_examples}).
As such, the surrounding pixels around a logo in any synthetic images
are completely determined by the background image, 
rather than the exemplar images. 
This is very different from \cite{eggert2015benefit}
where (1) pixel-level logo masks are extracted by tedious manual annotation 
and (2) the appearance of nearby pixels are biased largely towards the source images
thus may inevitably break the diversity of surrounding context 
(see Figure \ref{synth background compare} for example). 

\begin{table}  %
	\footnotesize
	\setlength{\tabcolsep}{0.07cm}
	\centering
	\caption{
		The popularity of TopLogo-10 brand logos as ranked by online media reports, among which not all brands
                were covered in every report. Smaller numbers indicate
                higher popularity. 		
	}
	\vskip 0.1cm
	\label{tab:logoselection}
	\begin{tabular}{c||c|c|c|c}
		\hline
		Criterion  & Clothing \cite{top10best}  & Luxury \cite{2015brandzluxury} & Fashion \cite{top10fashion} & Streetwear \cite{50streetwear} \\
		\hline \hline
		Adidas Classic  
		& 3 & - & - & -  \\
		\hline
		Helly Hansen
		& - & - & 1 & -  \\
		\hline
		Gucci
		& 5 & 3 & - & -  \\
		\hline
		Nike
		& 1  & -  & 8 & -   \\
		\hline
		Lacoste
		& 9 & - & - & -  \\
		\hline
		Chanel
		& -  & 4 & 7 & -  \\
		\hline
		Puma
		&  11 & - & - & -  \\
		\hline
		Micheal Kors
		& - & 9 & - & -  \\
		\hline
		Prada
		& - & 7 & - & -  \\
		\hline
		Supreme
		& -  & - & - & 2  \\
		\hline
	\end{tabular}
\end{table}

\vspace{0.2cm}
\noindent {\bf Logo Selection. }
We selected logos by considering:
{\bf (1)} Top popular and luxury brands of clothing/wearable
brandname logos based on recent online media reports. 
Specifically, 
	we particularly explored the clothing brand voting ranking \cite{top10best} 
	where the brands ``Nike", ``Adidas classic", ``Lacoste" and ``Puma" were selected; 
	and the luxury brand ranking \cite{2015brandzluxury}
	where ``Gucci", ``Chanel", ``Prada" and ``Michael Kors" were chosen. 
	Also, we picked the top jackets and sports brand 
	``Helly Hansen" \cite{top10fashion} and 
	one of the most popular street clothing brand ``Supreme" \cite{50streetwear}. 
	Table \ref{tab:logoselection} summarises the popularity ranking of these $10$ brand logos.
{\bf (2)} All $32$ logo classes in the popular FlickrLogo-32 dataset.
{\bf (3)} Common logos/brands in real-life, 
such as software, computer, website, food, university.
In total, we have $463$ logos from a wide range of products,
things and places in diverse context.
Example logos are shown in Figure \ref{fig:logo_examples}.


\vspace{0.2cm}
\noindent {\bf Context Images. } 
For context images, we used
the $6,000$ non-logo image subset of the FlickrLogo-32 dataset  \cite{romberg2011scalable}, 
obtained from the Flickr website by
query keywords ``building'', ``nature'', ``people'' and ``friends''.
These images (Figure \ref{fig:Synthetic_Image_Examples}) 
present highly diverse natural scene background in which 
logo may appear during synthesising training data.
Moreover, this ensures that all synthesised logo instances
are guaranteed to be labelled in the synthesised images, i.e., fully
annotated with no missing logos.

\subsection{Logo Exemplar Transformation}

\begin{figure}
	\centering
	\includegraphics[width=1\linewidth]{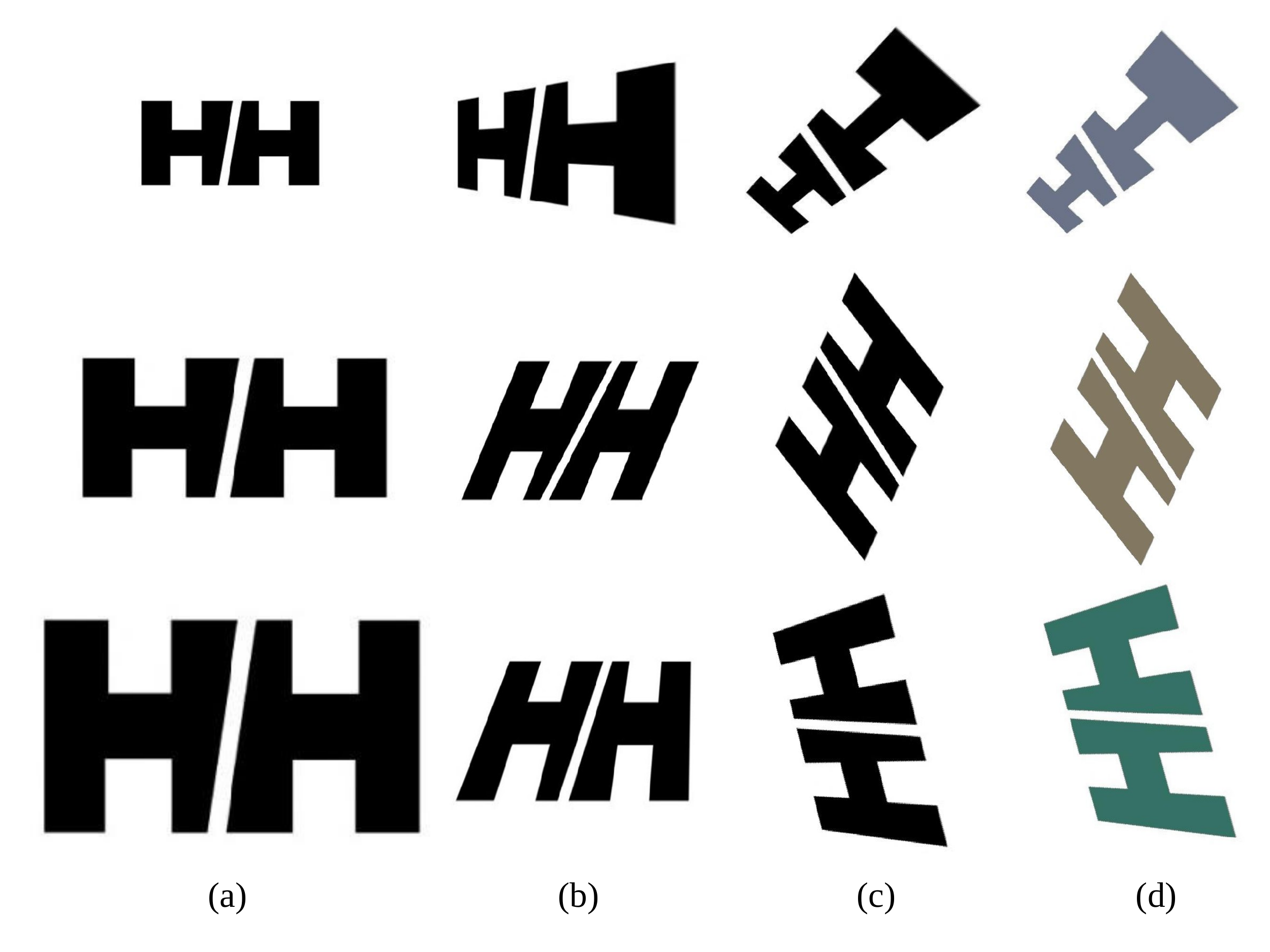}
	\caption{Illustration of logo exemplar transformations: 
		(a) scaling; 
		(b) shearing; 
		(c) rotation; 
		(d) colouring.
		We exploited these transformations jointly
for synthesising new training images.}
	\label{logosynthetic}
\end{figure}

In addition to context variations, we consider other sources of logo
variation are due to illumination change and geometric/perspective transforms.
These factors are approximated by warping logo exemplar images.
Formally, two independent types of transforms are deployed as follows \cite{eggert2015benefit}. 

\vspace{0.2cm}
\noindent {\bf (I) Geometric Transform. }
Suppose a logo image $\bm{I}$ is on a 3D plane, a general geometric
transform of the image is computed as:
\begin{equation}
\bm{I}^* = \bm{P} \bm{I} \bm{R}_{x} \bm{R}_{y} 
\label{synthetic_transform}
\end{equation}
where the matrix $\bm{P}$ defines 
a {\em scaling} (Figure \ref{logosynthetic} (a)) and 
{\em shearing} (Figure \ref{logosynthetic} (b))
projections;
$\bm{R}_{x}$ and $\bm{R}_{y}$ represent the
{\em rotation} (Figure \ref{logosynthetic} (c)) matrices for the corresponding axis respectively,
with the angles uniformly sampled from a range of [$0,360$]. 
	More precisely,
	we perform exemplar image geometric transforms in three steps
        as follows: 
	(i) Random scaling of the exemplar images (Figure \ref{logosynthetic} (a));
	(ii) Random shearing of the scaled images 
	(Figure \ref{logosynthetic} (b)); 
	(iii) Random rotation of the sheared images (Figure \ref{logosynthetic} (c)).
%

\vspace{0.2cm}
\noindent {\bf (II) Colouring Transform. }  
We also modify
the colour appearance of logo exemplar images for synthesising
illumination variations.
Specifically, we vary the pixel value in the RGB colour space
as
\begin{equation}
{c}^* = r {c}  
\label{synthetic_colourchange}
\end{equation}
where the scalar $c$ represents the pixel colour value in any channel and 
$r$ a random number sampled uniformly from [$0, 2$].
The range of $c^*$ is set to [$0,255$]
\footnote{One may come across zero-valued (pure black) pixels in
        clean logo exemplars. In this case, a multiplication operation based
        transform is invalid. Instead, we simply set the pixel value to 100 before performing colour variation
	(Figure \ref{logosynthetic}(d)).}.

\subsection{Synthesising Context Logo Images}
\label{SynData}

\begin{figure} 
	\centering
	\includegraphics[width=1\linewidth]{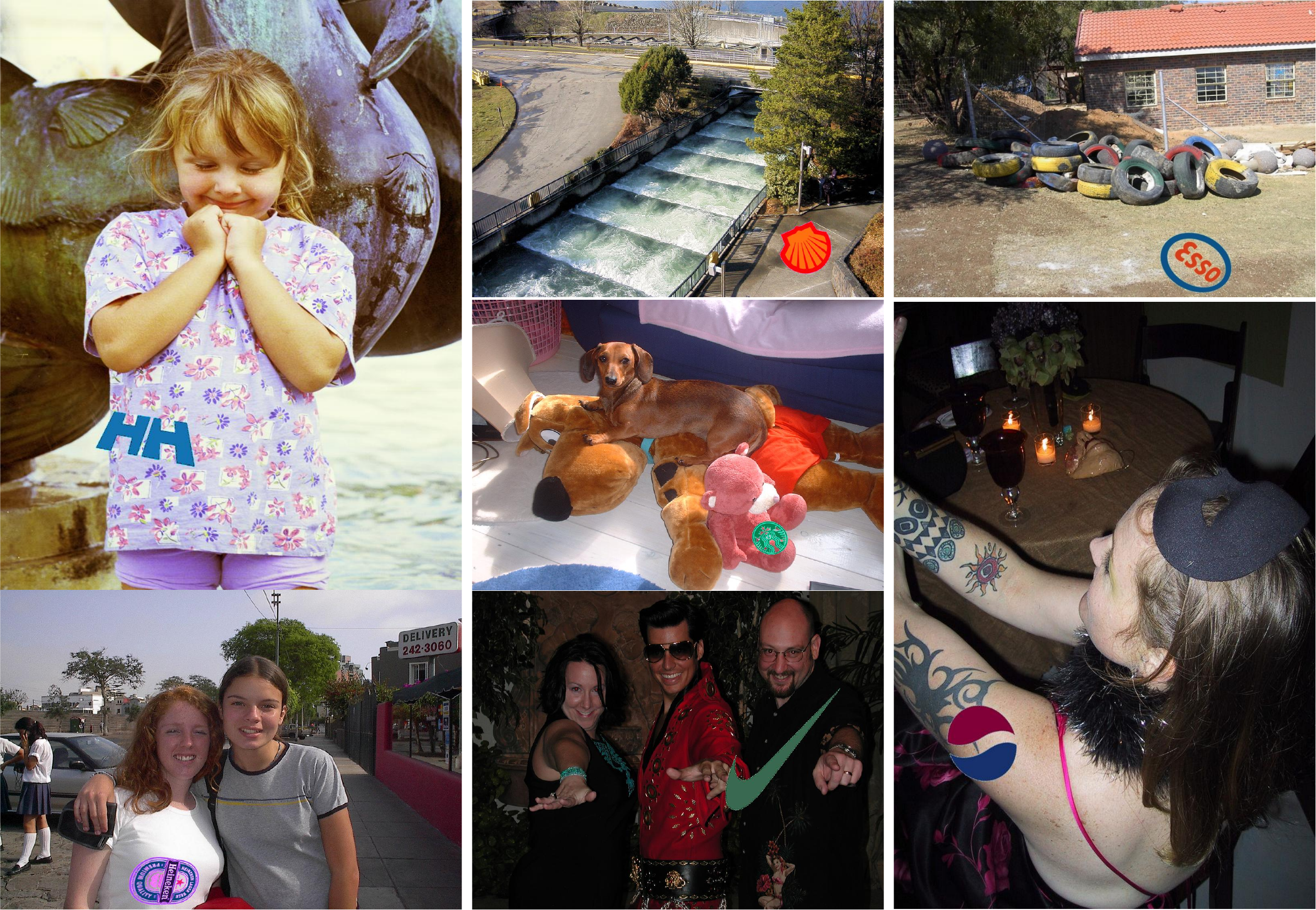}
	\caption{Examples of generated synthetic context images. Given
                    non-logo images from the FlickrLogo-32 dataset
          \cite{romberg2011scalable} as context images, these
          synthetic training images are fully annotated automatically
          without any logo instances unlabelled/missing.
	}
	\label{fig:Synthetic_Image_Examples}
\end{figure}

Given the logo exemplar image transforms described above,
we generate a number of variations for each logo,
and utilise them to synthesis logo images in context
by overlaying a transformed logo exemplar 
at a random location in non-logo context images.
This randomness in logo placement provides a large variety of
plausible visual scene context to enrich the synthesised logo training
images. 
%
%
For every logo class, we generate 100 synthetic logo images in context
through randomly selecting context images and applying geometric plus colouring transforms,
resulting in 46,300 synthetic context logo training images.
Examples of synthetic context logo images are shown in Figure \ref{fig:Synthetic_Image_Examples}.
\section{Experiments}\label{Experiments}

\begin{figure}
	\centering
	\includegraphics[width=1\linewidth]{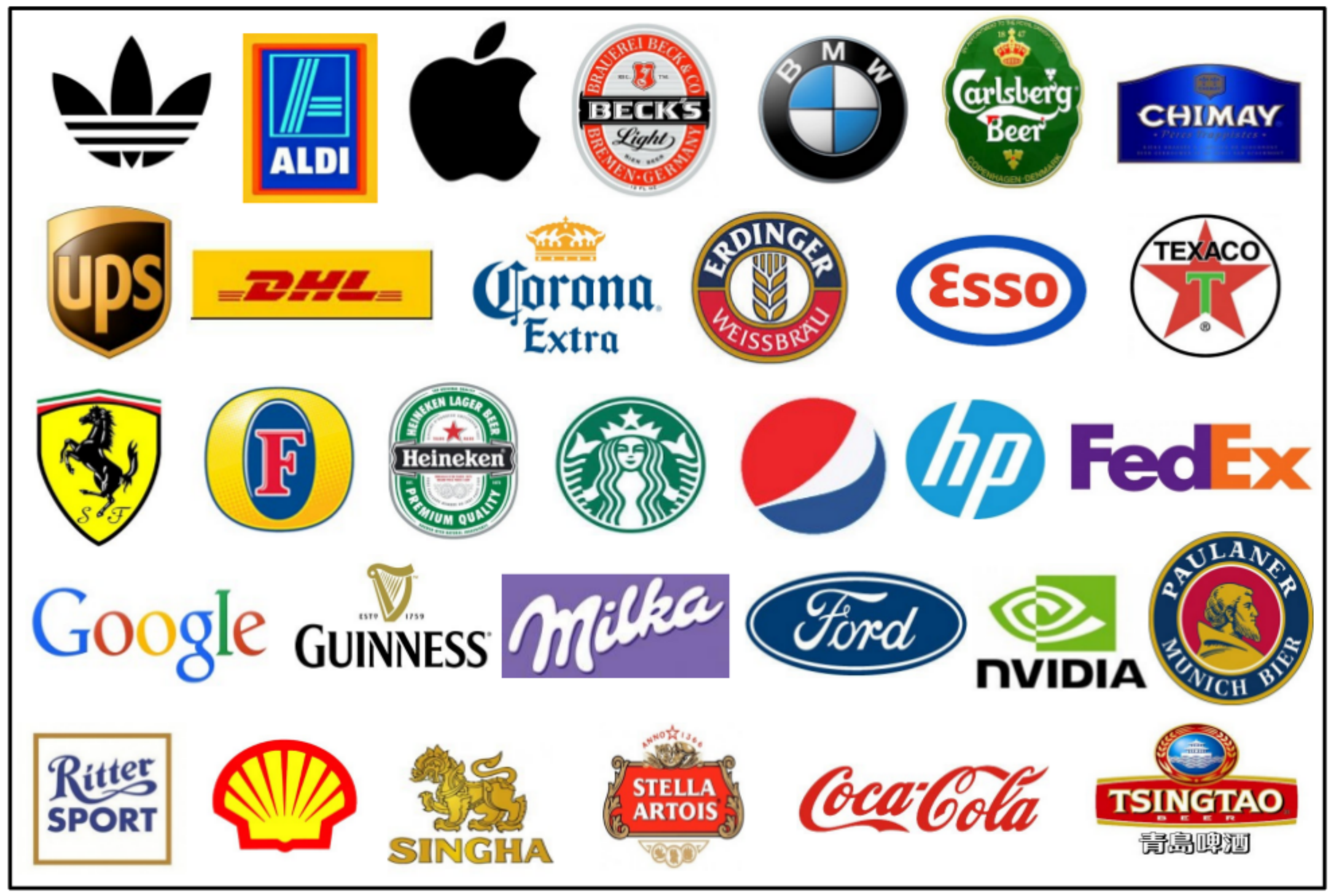}
	\includegraphics[width=1\linewidth]{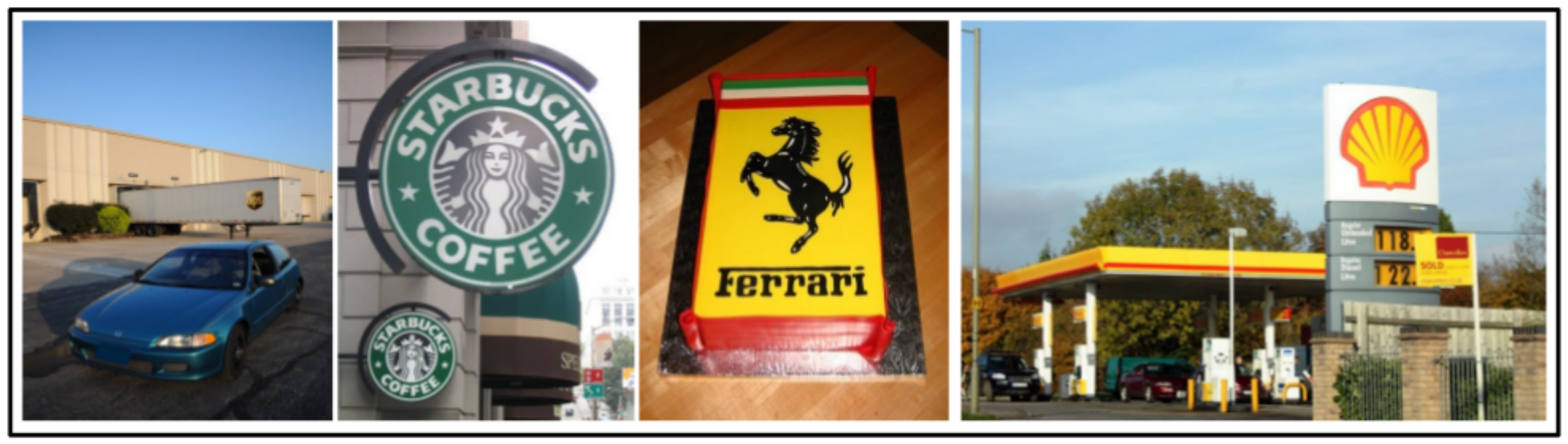}
	\caption{Exemplar images of 32 logo classes ({\bf Top}) 
		and test image examples ({\bf Bottom}) from the FlickrLogo-32 dataset \cite{romberg2011scalable}.}
	\label{flickr32_cleanlogo_example}
\end{figure}

\begin{figure}
	\centering
	\includegraphics[width=1\linewidth]{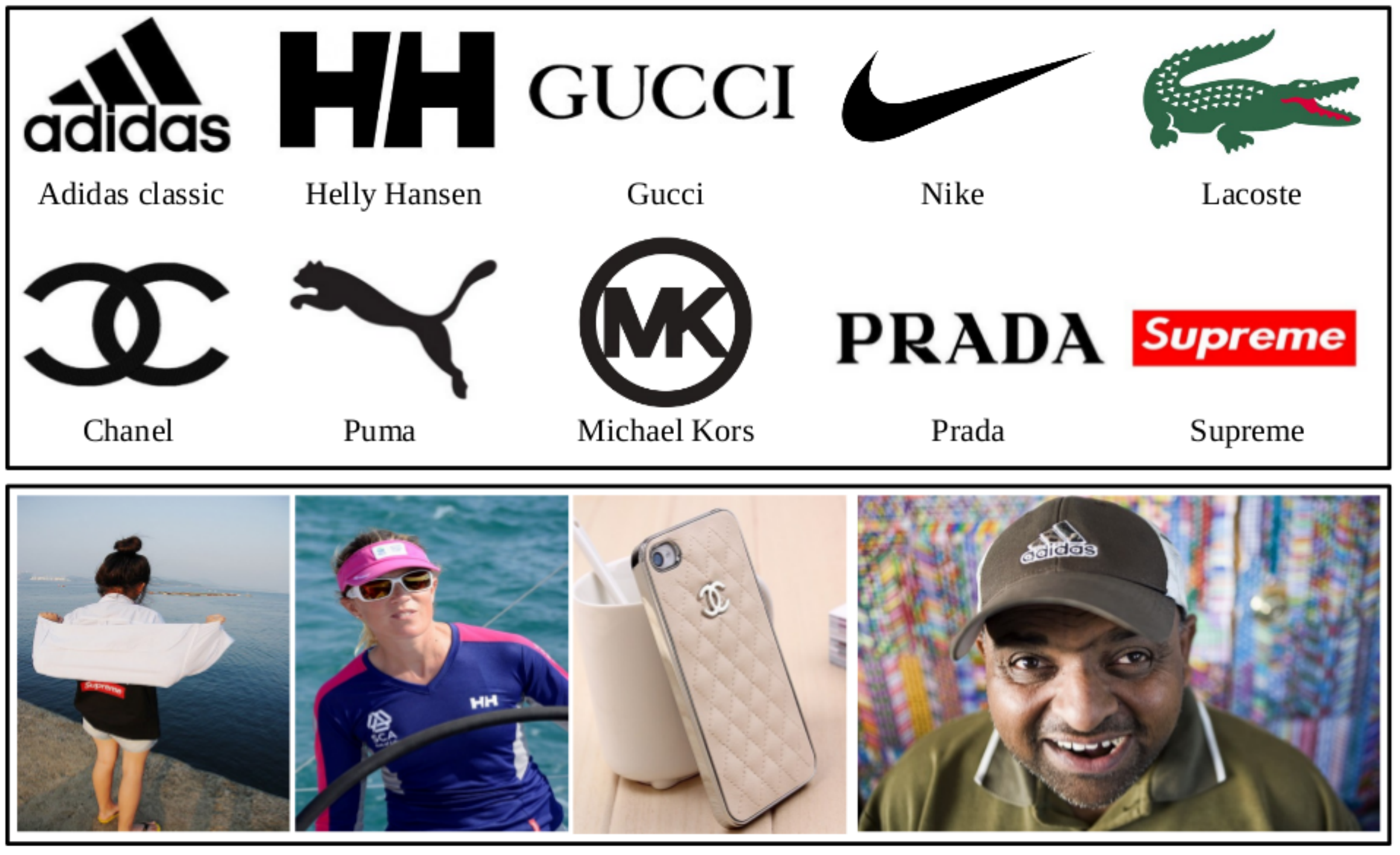}
	\caption{Exemplar images of top 10 logos ({\bf Top}) and test image examples ({\bf Bottom}) from our TopLogo-10 dataset.}
	\label{fig:toplogo}
\end{figure}

\subsection{Datasets}
\label{sec:datasets}

Two logo detection datasets were utilised for evaluation:
{FlickrLogo-32} \cite{romberg2011scalable} and 
TopLogo-10, newly introduced in this work.

\vspace{0.1cm}
\noindent {\em FlickrLogo-32}.
This is the most popular logo detection dataset containing $32$
different logos (Table \ref{tab:dataset}). Examples of logo
exemplars and test images are shown in Figure \ref{flickr32_cleanlogo_example}.

\vspace{0.1cm}
\noindent {\em TopLogo-10}.
From 463 logos, we selected Top-$10$ clothing/wearable
brandname logos by popularity/luxury, and constructed a logo dataset
with manual labelling. We call this new dataset {\em TopLogo-$10$}.
Specifically, there are ten logo classes: ``Nike'',``Chanel'', ``Lacoste'', ``Gucci'', ``Helly Hansen'',
``Adidas Classic'', ``Puma'', ``Michael Kors'', ``Prada'', ``Supreme'',
with various degrees of composition complexity in the logos.
For each logo class, $70$ images are included,
each with fully manually labelled logo bounding boxes.
The exemplars and examples of test images are shown in Figure \ref{fig:toplogo}.
These logo instances may appear in a variety of context, e.g.,
shoes, hats, shower gels, wallets, phone covers, lipsticks,
eye glasses, spray, jackets, T-shirts, peaked caps, and sign boards.
Moreover, logo instances in TopLogo-10 have varying sizes as in natural
images, therefore imposes significant detection challenges from small
sized logos.
TopLogo-10 represents some of the common and natural logo existence
scenarios and provides realistic challenges for logo detection.

For each of the two datasets, we divided randomly all the images into two parts:
(1) one for training, with 10 images per logo class;
(2) one for testing, with the remaining images.
Given such a small number of labelled images, it is very challenging
to train deep logo models with millions of parameters.

\subsection{Baseline Methods}
\label{sec:baseline}
For evaluating the effectiveness of our synthetic context logo images 
for learning a deep logo detector, we utilised the Faster R-CNN \cite{ren2015faster} as 
our logo detector. Other object detectors
\cite{redmon2015you,liu2015ssd} are also available but this is
independent from our proposed method.
We trained a Faster R-CNN detector by the following different processes
and comparatively evaluated their logo detection performances.
{\bf (1)} RealImg: 
Only labelled real training image are used
for model training.
{\bf (2)} SynImg-$x$Cls: 
The synthetic labelled training data from $x$ ($x=32$ for
FlickrLogo-32 and $x=10$ for TopLogo-10) target logo classes are used
for model training;
{\bf (3)} SynImg-$463$Cls: 
The synthetic labelled training data from all $463$ logo classes are
used for model training;
{\bf (4)} SynImg-$x$Cls+RealImg: 
We first utilise the synthetic training data
from the target logo classes to pre-train a Faster R-CNN,
then fine-tune the model using labelled real training data;
this is a staged curriculum learning model training scheme
\cite{bengio2009curriculum};
{\bf (5)} SynImg-$463$Cls+RealImg: 
Similar to {SynImg-$x$Cls+RealImg} but with a difference that
all synthetic training data are used for pre-training the detector;
{\bf (6)} SynImg-$463$Cls+RealImg (Fusion): 
Similar to {SynImg-$463$Cls+RealImg} but with a difference that
the model is trained in a single step using the fusion of synthetic
and realistic training data, other than the sequential curriculum learning.  

\subsection{Evaluation Metrics}
\label{sec:metric}
All models are trained on a training set and evaluated on
a separate independent testing set. 
The logo detection accuracy is measured by Average Precision (AP) for each
individual logo class and mean Average Precision (mAP)
over all logo classes \cite{ren2015faster}.

\subsection{Implementation Details}
For learning a logo Faster R-CNN detector, 
we set the learning rate as $0.0001$ on 
either synthetic or real training data.
The learning iteration is set to $40,000$.
For all the models, we (1) pre-trained a Faster R-CNN with the training data of the ImageNet-1K object classification dataset \cite{russakovsky2014imagenet}
for parameter initialisation \cite{simonyan2014very},
and (2) then further fine-tuned the model on PASCAL VOC non-logo object detection images \cite{everingham2010pascal}.


\subsection{Evaluations}
\label{sec:eval}

\begin{table*} 
	\footnotesize
	\centering
	\setlength{\tabcolsep}{0.3cm}
	\caption{
		Evaluating different methods on the FlickrLogo-32 dataset \cite{romberg2011scalable}.
		RealImg: Realistic Image; SynImg: Synthetic Image.
	}
	\vskip 0.1cm
	\label{tab:flickr32}
	\begin{tabular}{c||c||cccccccc||c}
		\hline
		 & Setting: & Adidas & Aldi & Apple & Becks & BMW & Carls & Chim & Coke & \\
		Method & Training/Test & Corona & DHL & Erdi & Esso & Fedex & Ferra & Ford & Fost &  mAP \\
		& Image Split & Google & Guin & Hein & HP & Milka & Nvid & Paul & Pepsi &  \\
		& (per logo class) & Ritt & Shell & Sing & Starb & Stel & Texa & Tsin & Ups &  \\
		\hline \hline
		& & 23.7 & 57.5 & 63.0 & 69.6 & 63.7 & 50.6 & 55.2 & 26.8 & \\
		RealImg & Training: 10 RealImg & 79.0 & 25.8 & 61.2 & 44.2 & 45.9 & 80.6 & 64.3 & 43.2 & \\
		&  & 47.7 & 58.2 & 61.8 & 21.3 & 19.4 & 17.4 & 48.2 & 17.8 & \\
		& Test: 60 RealImg & 34.8 & {\bf 45.8} & 71.8 & 70.2 & 79.6 & 56.7 & 56.9 & 52.2 & 50.4\\
		\hline

		& & 9.4 & 47.3 & 9.6 & 70.3 & 39.9 & 28.3 & 15.8 & 21.7 & \\
		SynImg-$32$Cls & Training: 100 SynImg & 6.1 & 11.1 & 4.1 & 44.7 & 22.9 & 60.9 & 43.6 & 28.8 & \\
		&  & 23.0 & 16.7 & 43.1 & 9.9 & 4.6 & 1.1 & 38.1 & 9.7 & \\
		& Test: 60 RealImg & 22.7 & 38.3 & 15.5 & 65.6 & 28.7 & 55.1 & 27.4 & 20.1 & 27.6\\
		\hline
		&  & 9.2 & 24.8 & 4.5 & 30.5 & 24.1 & 15.3 & 2.4 & 20.9 & \\
		SynImg-$463$Cls  & Training: 100 SynImg & 0.4 & 3.8 & 4.8 & 48.7 & 20.5 & 45.2 & 29.0 & 24.5 & \\
		&  & 13.0 & 1.1 & 27.8 & 10.4 & 1.8 & 7.2 & 26.1 & 10.8 & \\
		& Test: 60 RealImg & 18.6 & 38.6 & 0.7 & 49.3 & 28.5 & 61.1 & 23.7 & 17.5 & 20.5\\
		\hline
		&  & {\bf 26.8} & 63.7 & 65.8 & 72.7 & {\bf 81.3} & 52.7 & {\bf 63.6} & 30.0 & \\
		SynImg-$32$Cls & Training: 100 SynImg & 76.0 & 31.5 & 63.0 & 52.2 & {\bf 54.3} & {\bf 90.0} & {\bf 84.0} & 46.6 & \\
		 + RealImg & +10 RealImg & 58.0 & 52.6 & 65.2 & 23.2 & {\bf 24.0} & 12.5 & 54.1 & {\bf 23.6} & \\
		& Test: 60 RealImg & 37.9 & 45.6 & {\bf 75.0} & {\bf 73.8} & 79.0 & {\bf 64.2} & 57.4 & {\bf 54.4} & 54.8\\
		\hline
		&  & 22.6 & {\bf 66.6} & {\bf 72.0} & {\bf 73.2} & 78.7 & {\bf 53.3} & 58.0 & {\bf 31.2} & \\
		SynImg-$463$Cls   & Training: 100 SynImg & {\bf 82.7} & {\bf 33.7} & {\bf 67.2} & {\bf 53.5} & 50.8 & 85.6 
		& 72.4 & {\bf 51.3} & \\
		+ RealImg & + 10 RealImg & {\bf 59.6} & {\bf 67.7} & {\bf 69.6} & {\bf 28.1} & 21.9 & {\bf 17.4} & {\bf 59.6} & 21.8 & \\
		& Test: 60 RealImg & {\bf 42.7} & 45.5 & 74.0 & 72.3 & {\bf 83.1} & 63.6 & {\bf 60.2} & 49.3 & {\bf 55.9}\\
		\hline 
		&  & 10.4 & 45.2 & 3.7 & 63.0 & 41.6 & 27.8 & 9.6 & 22.8 &  \\
		SynImg-$463$Cls & Training: 100 SynImg & 10.6  & 14.0 & 5.5 & 56.9 & 28.4 & 62.9 & 48.2 & 53.6 & \\
		+ RealImg & + 10 RealImg & 44.1 & 21.1 & 47.1 & 10.6 & 6.2 & 5.2 & 52.9 & 15.0 & \\
		(Fusion) & Test: 60 RealImg & 37.7 & 36.8 & 4.5 & 59.4 & 25.1 & 67.4 & 27.7 & 22.4 & 30.9\\
		\hline \hline
		& & \bf 68.1 & \bf 79.1 & 84.5 & 72.3 & \bf 86.4 & \bf 68.0 & \bf 78.0 & \bf 73.3 & \\
		{RealImg} & Training: 40 RealImg & 90.9 & \bf 77.4 & \bf 90.9 & 88.6 & \bf 71.1 & \bf 91.0 & \bf 98.3 & \bf 86.2 & \\
		&  & \bf 98.0 & \bf 90.7 & \bf 81.3 & \bf 67.0 & \bf 54.5 & \bf 64.0 & 90.9 & \bf 59.6 & \\
		& Test: 30 RealImg & \bf 81.0 & 57.3 & \bf 97.9 & \bf 99.5 & \bf 86.7 & \bf 90.4 & \bf 87.5 & \bf 85.8 &\bf 81.1 \\
		\hline
		&  & 61.6 & 67.2 & \bf 84.9 & \bf 72.5 & 70.0 & 49.6 & 71.9 & 33.0 & \\
		Deep Logo \cite{iandola2015deeplogo} 
		& Training: 40 RealImg & \bf 92.9 & 53.5 & 80.1 & \bf 88.8 & 61.3 & 90.0 & 84.2 & 79.7 & \\
		&  & 85.2 & 89.4 & 57.8 & - & 34.6 & 50.3 & \bf 98.6 & 34.2 & \\
		  & Test: 30 RealImg & 63.0 & \bf 57.4 & 94.2 & 95.9 & 82.2 & 87.4 & 84.3 & 81.5 & 74.4 \\
		\hline
	    &  & - & - & - & - & - & - & - & - & \\
		BD-FRCN-M \cite{oliveira2016automatic}  
		& Training: 40 RealImg & - &  -& - & - & - & - &-  & - & \\
		&  & - & - & - & - & - &  -& - & - & \\
		& Test: 30 RealImg & - & - & - & - & - & - & - & - & 73.5\\
		\hline
		
	\end{tabular}
\end{table*}

\begin{figure} 
	\centering
	\includegraphics[width=1\linewidth]{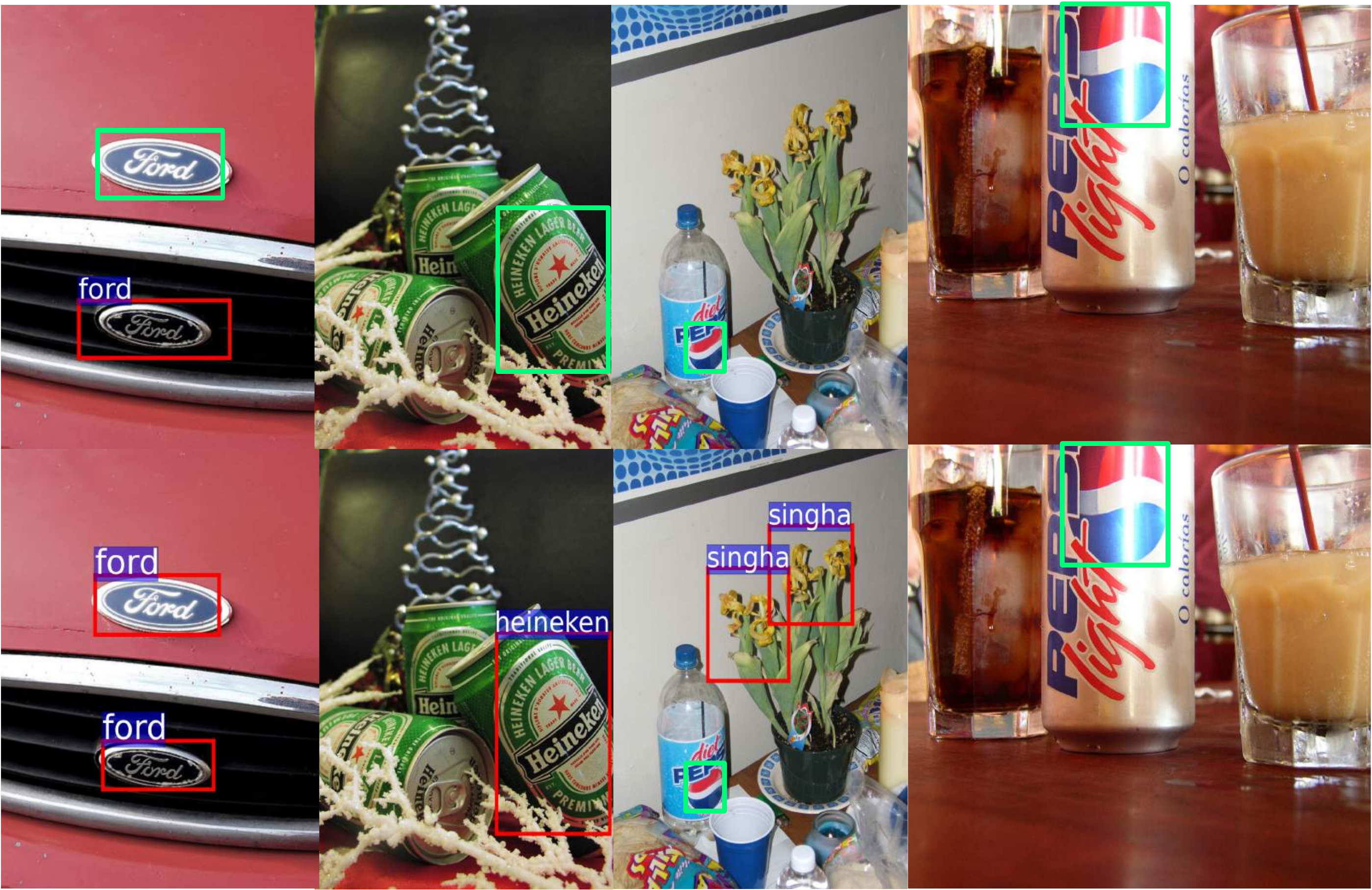}
	\caption{Qualitative evaluation on FlickrLogo-32 \cite{romberg2011scalable}. 
		{\bf First row}: detection results by ``RealImg'';
		{\bf Second row}: detection results by ``SynImg-$463$Cls + RealImg''.
		Logo detections are indicated with red boxes.
		Ground truth is indicated by green boxes.
	}
	\vspace{-0.3cm}
	\label{fig:qualitativeFlickr32}
\end{figure}

\noindent {\bf Evaluation on FlickrLogo-32. }
We performed logo detection on the FlickrLogo-32 logo images \cite{romberg2011scalable}.
The results of different methods are shown in Table \ref{tab:flickr32}.
It is evident that logo detection performance by Faster R-CNN can be 
largely improved by expanding the training data with 
the synthetic context logo images generated
with our proposed method. 
For example, the combination of full synthetic and realistic training data
(i.e., SynImg-$463$Cls + RealImg) 
can boost the detection accuracy from $50.4\%$ (by RealImg) to $55.9\%$,
with $5.5\%$ increase in absolute mAP or $10.9\%$ relative improvement. 
Importantly, such performance gain is achieved {\em without} any additional exhaustive labelling but
by automatically enriching the context variations in the training data.

Specifically, we draw these following observations.
{\em Firstly}, by using merely 10 training images per logo (i.e., RealImg),
Faster R-CNN is already able to achieve fairly good detection results ($50.4\%$ in mAP).
This suggests the great capability of deep models partially due to 
the good parameter initialisation on ImageNet and PASCAL VOC images, 
confirming the similar findings elsewhere \cite{hoffman2013one,bengio2012deep,bengio2011deep}.
{\em Secondly},
when using our synthetic training images alone 
(i.e., SynImg-$32$Cls and SynImg-$463$Cls), 
the model performance on test data is much inferior 
than using sparse realistic training data.
The potential reasons is that,
there may exist great discrepancy between realistic and synthetic training images, 
also known as the domain drift problem \cite{torralba2011unbiased,nguyen2015dash}, 
i.e., a trained model may degrade significantly in performance when deployed to a new domain
with much disparity involved as compared to the training data.
{\em Thirdly}, 
it is observed interestingly that synthetic training data from non-target logo classes
may even hurt the model generalisation, 
when comparing the mAP result between SynImg-$32$Cls ($27.6\%$) and SynImg-$463$Cls ($20.5\%$).
This may be due to the distracting effects introduced during detector optimisation
on a large number of ($431$) non-target logos and thus making the resulting model
less effective towards target logos in model deployment.
{\em Fourthly}, it is also found that model pre-training on the full synthetic images
turns out to be superior than on those of target logo classes alone.
This suggests that more generic model pre-training may produce better initialisation 
for further incremental model adaptation on sparse real training data.

\begin{table*} 
	\footnotesize
	\centering
	\setlength{\tabcolsep}{0.22cm}
	\caption{
		Evaluating different methods on our TopLogo-10 logo dataset. 
	}
	\vskip 0.1cm
	\label{tab:toplogo10}
	\begin{tabular}{c||c|c|c|c|c|c|c|c|c|c||c}
		\hline
		Method & Adidas & Chanel & Gucci & HH & Lacoste & MK & Nike & Prada & Puma & Supreme & mAP \\
		\hline \hline
		RealImg & 28.6 & 32.9 & 32.8 & 33.9 & 47.1 & 40.4 & 0.5 & 15.0 & 9.5 & 44.4 & 28.5\\
		\hline
		SynImg-$10$Cls & 7.1 & 9.2 & 3.0 & 0.0 & 10.9 & 13.5 & 0.1 & 0.2 & 9.1 & 20.1 & 7.3\\
		\hline
		{SynImg-$463$Cls} & 14.1 & 4.7 & 9.1 & 0.4 & 18.3 & 22.9 & 3.0 & 0.2 & 4.0 & 25.1 & 10.2\\
		\hline
		
		SynImg-$10$Cls + RealImg & 51.9 & {\bf 44.8} & 41.1 & {\bf 38.1} & {\bf 53.3} & 52.5 & 11.8 & 28.9 & 18.4 & 63.6 & 40.4\\
		\hline
		SynImg-$463$Cls + RealImg & {\bf 52.7} & 39.9 & {\bf 49.7} & 36.5 & 48.4 & {\bf 62.7} & {\bf 14.8} & {\bf 29.8} & {\bf 18.6} & {\bf 64.6} & {\bf 41.8}\\
		\hline
		
		
		SynImg-$463$Cls + RealImg (Fusion)
		& 25.4 & 11.8 & 4.7 & 0.6 & 17.9 & 44.2 & 1.6 & 0.9 & 15.5 & 40.3 & 16.3\\
		\hline
	\end{tabular}
\end{table*}

\begin{figure*} 
	\centering
	\includegraphics[height=0.4\linewidth]{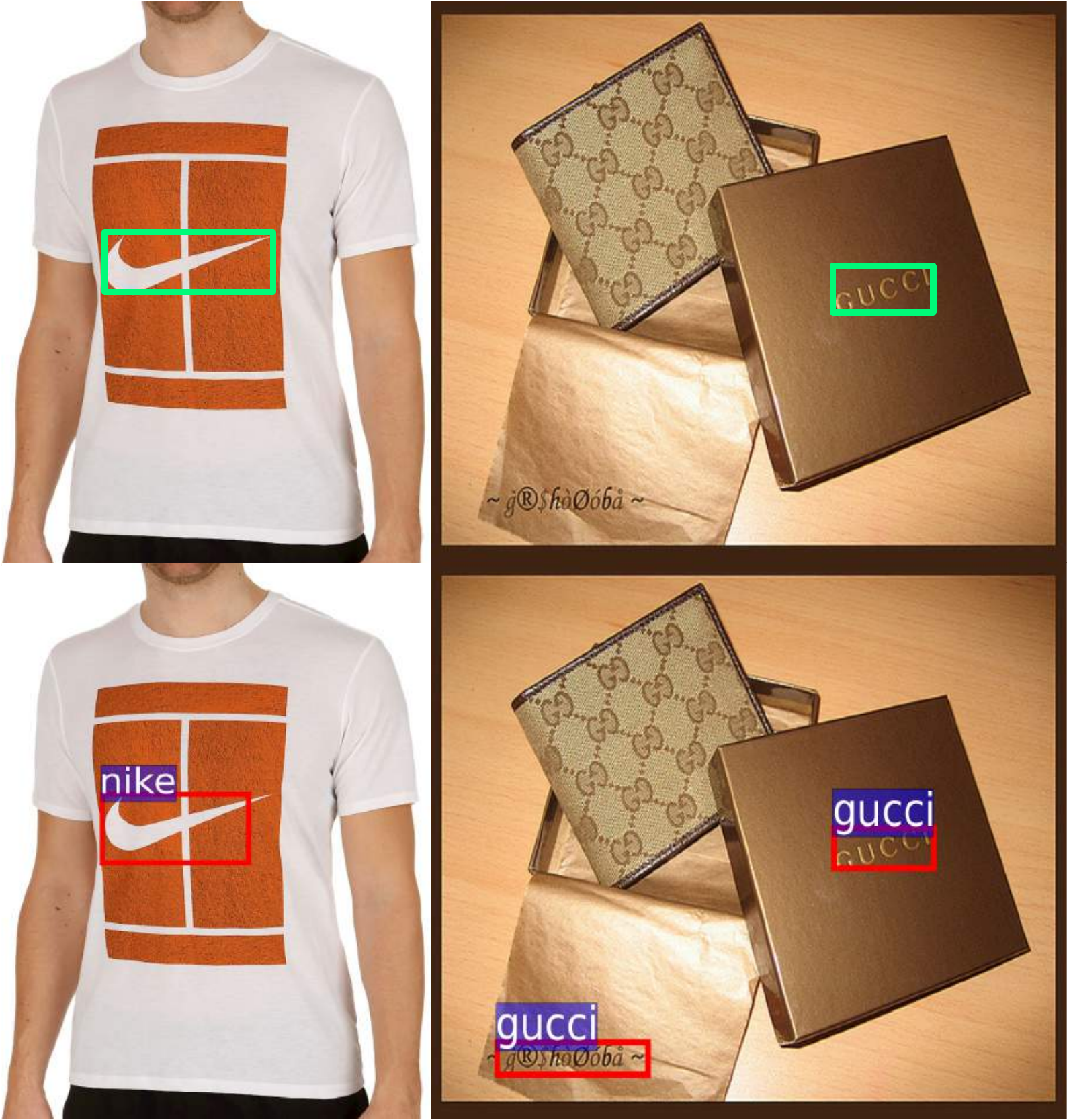}
	\includegraphics[height=0.4\linewidth]{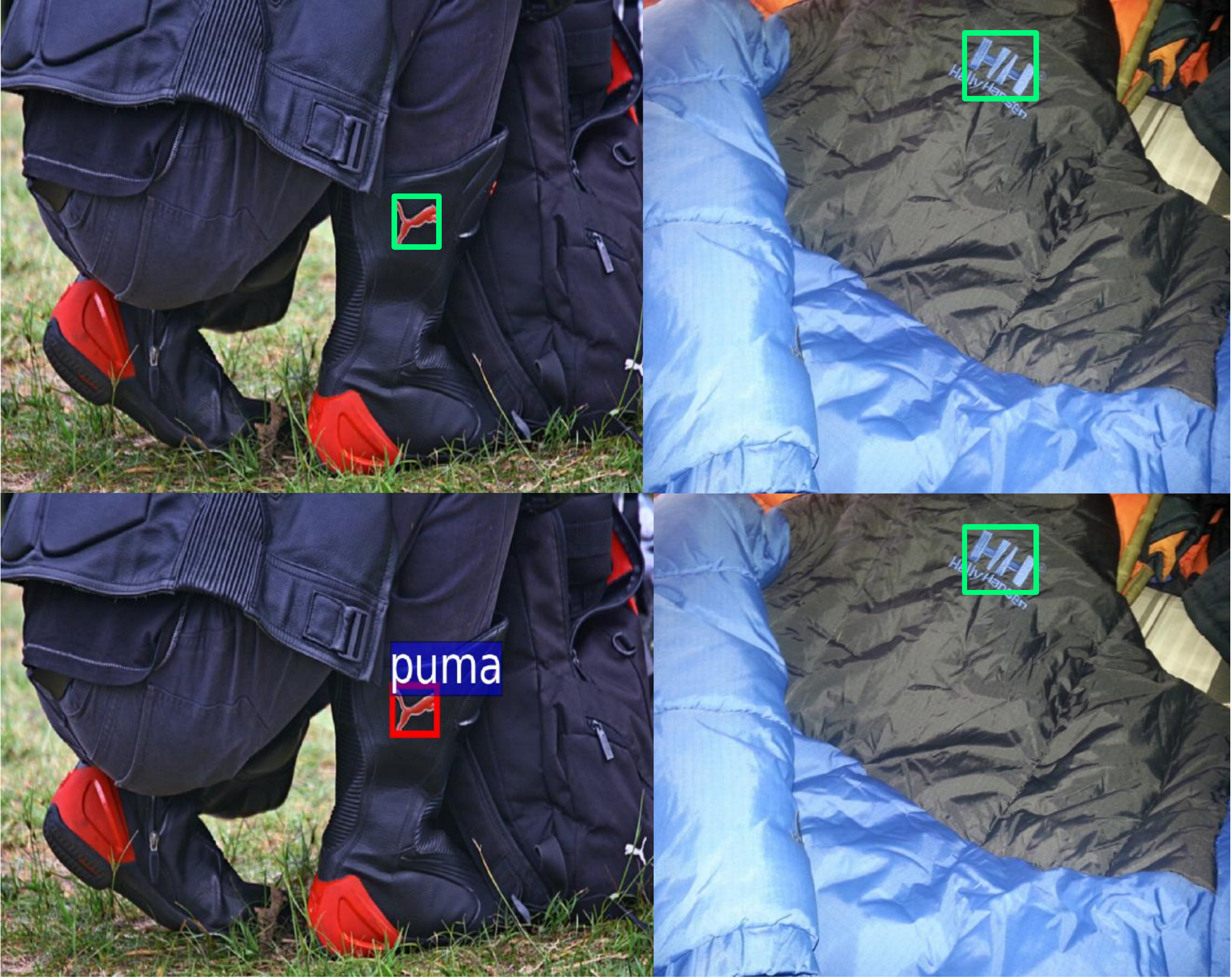}
	\caption{Qualitative evaluations on our TopLogo-10 logo dataset.
		{\bf First row}: detection results by ``RealImg'';
		{\bf Second row}: detection results by ``SynImg-$463$Cls + RealImg''.
		Logo detections are indicated with red boxes.
		Ground truth is indicated by green boxes. 
	}
	\label{fig:qualitativeToplogo10}
\end{figure*}

	In addition to comparative evaluations under the same setting, 
	we also compare the reported results from the state-of-the-art
        methods  \cite{iandola2015deeplogo,oliveira2016automatic}. 
	It can be seen (last three rows in Table \ref{tab:flickr32})
	that the best alternative DeepLogo surpasses our method (SynImg-463Cls + RealImg):
	$74.4\%$ vs. $55.9\%$ in mAP.
	However, it should also be pointed out that our model was
        trained on much less ($25\%$) real training images. 
	To provide a more comparable evaluation, we exploited the
        Faster R-CNN model as an alternative to DeepLogo,
	since DeepLogo code is not released publicly, and trained a
        new model using same training data split as in DeepLogo. 
	Our results on the FlickrLogo-32 dataset under the DeepLogo
        $40/30$ image split by our Faster RCNN based logo detector is
        $81.1\%$ (the third last row in Table \ref{tab:flickr32}), 
        significantly better than DeepLogo’s $74.4\%$. 
	Moreover, under our $10/60$ split, a Faster R-CNN based logo
        detector (RealImg in Table \ref{tab:flickr32})  
	yields $50.4\%$ in mAP vs. $55.9\%$ by our method SynImg-463Cls + RealImg. 
	Taken both into account, it suggests that the proposed method
        is more likely to outperform the DeepLogo model if compared
        under the same $10/60$ split setting. Given that this work
        focuses on exploring synthetic training data for deep learning
        logo detection regardless the detection model, any other detector \cite{liu2015ssd} 
	can be readily integrated into our framework.

	We further evaluated the effect of the curriculum learning strategy in detection performance.
	To that end, we can compare SynImg-$463$Cls + RealImg and SynImg-$463$Cls + RealImg (Fusion)
	in Table \ref{tab:flickr32}.
	It is evident that by blindly learning a Faster R-CNN logo
        detector on the fusion (combined)
	of real and synthetic training images, the model
        generalisation is significantly degraded, likely due to severe bias towards the synthetic data. 
	This is because that the synthetic images dominate in numbers
        in combined training data, resulting in that the trained model suffers from
	the domain shift problem. This result validates the efficacy
        of the proposed curriculum learning method in learning a more robust 
	logo detection model when given heterogeneous training data sources. 	  

Lastly, we carried out qualitative evaluations
by examining the effect of synthetic context logo images 
on the detector performance. 
Figure \ref{fig:qualitativeFlickr32} shows that context
variations/diversity is effective for improving deep model learning
resulting in more discriminative logo features therefore less missing
logo detections -- less false negatives (see $1^\text{st}$ and $2^\text{nd}$ columns).
It is also evident that both models have difficulties in detecting small logos ($3^\text{rd}$ column)
and logos under extreme lighting conditions ($4^\text{th}$ column).
Also, the model trained by a single stage with full synthetic and real training data is likely to
generate more false positive detections ($3^\text{rd}$ column), 
likely due to over-diversity (aka noise) introduced by the synthetic context.

\begin{table*} 
	\footnotesize
	\centering
	\setlength{\tabcolsep}{0.28cm}
	\caption{
		Evaluating the effect of different synthetic context.
		Dataset: TopLogo-10. 
	}
	\vskip 0.1cm
	\label{tab:eval_context}
	\begin{tabular}{c||c|c|c|c|c|c|c|c|c|c||c}
		\hline
		Logo Name & Adidas & Chanel & Gucci & HH & Lacoste & MK & Nike & Prada & Puma & Supreme & mAP \\
		\hline \hline
		{\bf Scene Context} & {\bf 7.1} & {\bf 9.2} & {\bf 3.0} & 0.0 & {\bf 10.9} & {\bf 13.5} & {\bf 0.1} & {\bf 0.2} & {\bf 9.1} & {\bf 20.1} & {\bf 7.3}\\

		Clean Context & 2.8 & 2.0 & 0.0 &\bf 4.5 & 0.5 & 6.0 & \bf 0.1 & 0.0 &\bf 9.1 & 10.2 & 3.5\\
		\hline

		{\bf Scene Context} + RealImg  & {\bf 51.9} & {\bf 44.8} & {\bf 41.1} & {\bf 38.1} & {\bf 53.3} & 52.5 & {\bf 11.8} & 28.9 & {\bf 18.4} & 63.6 & {\bf 40.4}\\
		Clean Context + RealImg 
		& 48.7 & 37.3 & 38.1 & 33.9 & 46.4 & {\bf 61.7} & 2.5 & {\bf 31.0} & 12.4 & {\bf 67.4} & 37.9\\
		No Context + RealImg & 28.6 & 32.9 & 32.8 & 33.9 & 47.1 & 40.4 & 0.5 & 15.0 & 9.5 & 44.4 & 28.5\\
		\hline
	\end{tabular}
\end{table*}

%

\begin{table*} [th!] %
	\footnotesize
	\centering
	\setlength{\tabcolsep}{0.34cm}
	\caption{
		Evaluating the effect of individual logo transformations. 
		Dataset: TopLogo-10.
	}
	\vskip 0.1cm
	\label{tab:eval synth data different transformation}
	\begin{tabular}{c||c|c|c|c|c|c|c|c|c|c||c}
		\hline
		Logo Name & Adidas & Chanel & Gucci & HH & Lacoste & MK & Nike & Prada & Puma & Supreme & mAP \\
		\hline \hline
		No Colouring & \bf 15.4 & 4.6 & 7.6 & 0.0 & 9.1 & 0.5 & \bf 9.1 & 4.5 & \bf 9.1 &  9.1 & 6.9\\
		+ RealImg  & 52.9 & 40.5 & \bf 46.4 & 27.7 & 53.2 & 48.0 & \bf 12.6 & 25.9 & 17.4 & \bf 66.6 & 39.1\\
		\hline
		No Rotation & 13.6 & \bf 9.2 & \bf 11.5 & \bf 3.0 & 13.6 & 10.8 & 0.3 & \bf 9.1 & \bf 9.1 & \bf 23.6 & \bf 10.4\\
		+ RealImg  & 56.7 & 37.5 & 44.7 & 30.7 & \bf 55.1 & 55.4 & 4.2 & \bf 35.3 & \bf 24.5 & 61.4 & \bf 40.5\\
		\hline
		No Scaling & 4.6 & 4.6 & 3.0 & 2.3 & 9.1 & 1.2 & 0.0 & 0.0 & \bf 9.1 & 12.0 & 4.6\\
		+ RealImg  & 55.7 & 35.3 & 46.3 & 31.2 & 54.7 & 49.0 & 10.6 & 30.6 & 13.2 & 61.6 & 38.8\\
		\hline
		No Shearing & 3.9 & 1.7 & 9.1 & 0.0 & 9.1 & 3.5 & 0.0 & 0.0 & \bf 9.1 & 12.1 & 4.9\\
		+ RealImg  & \bf 58.8 & 34.7 & 44.2 & 35.3 & 47.9 & \bf 55.7 & 6.0 & 24.0 & 16.2 & 60.9 & 38.4\\
		\hline 
		\hline
		{\bf Full} & 7.1 & \bf 9.2 & 3.0 & 0.0 & \bf 10.9 & \bf 13.5 & 0.1 & 0.2 & \bf 9.1 & 20.1 & 7.3\\
		+ RealImg 
		& 51.9 & {\bf 44.8} & 41.1 & {\bf 38.1} & 53.3 & 52.5 & 11.8 & 28.9 & 18.4 & 63.6 & 40.4\\
		\hline
	\end{tabular}
\end{table*}

\vspace{0.2cm}
\noindent {\bf Evaluation on TopLogo-10. }
We evaluated Faster R-CNN detectors trained with different methods on our 
TopLogo-10 dataset.
We present the detection results in Table \ref{tab:toplogo10}.

By exploiting our synthetic 
context logo training images,
the logo detection performance can be improved more significantly
than on FlickrLogo-32, e.g., from $28.5\%$ (by RealImg) to $41.8$ (by SynImg-$463$Cls + RealImg)
with $13.3\%$/$46.7\%$ absolute/relative boost in mAP,
as compared to $5.5\%/10.9\%$ on FlickrLogo-32.
This further suggests the effectiveness of our synthetic context 
driven training data expansion method for logo detection,
particularly for the practical and more challenging clothing brand logos.
Mostly, we observed similar phenomenons as those on FlickrLogo-32
but one difference that SynImg-$463$Cls outperforms considerably SynImg-$10$Cls.
The possible reason is that, 
the generic semantic context learned from a large number of logo classes becomes more 
indicative and useful since
the background of clothing logos tends to be more clutter than those from FlickrLogo-32.
Similarly, we show some qualitative detection examples in Figure \ref{fig:qualitativeToplogo10}.

\vspace{0.2cm}
\noindent {\bf Further Analysis.} 
	To give more insight, we performed further experiments with SynImg-$10$Cls on the TopLogo-10 dataset.

\vspace{0.1cm}
\noindent 
{\em The Effect of Different Synthetic Context.}
	We specifically evaluated the impact of context on model learning. 
	To that end, we introduce a new type of context - clean black context 
	(``Clean Context'') as an alternative to the natural scene
        context (``Scene Context'') used early in our SCL model.
	Table \ref{tab:eval_context} shows when only synthetic
        training images were used in model training, the ``Scene
        Context'' is much more superior than the ``Clean Context'' for
        model training.
	This advantage remains when real training images were exploited for model adaptation.
	Interestingly, we also found that the ``Clean Context'' is able to improve logo detection 
	performance, as revealed by comparison to the results without
        using any synthetic context.
	This further suggests that synthetic context based 
	data expansion is an effective strategy for addressing the training data scarcity challenge.

\vspace{0.1cm}
\noindent 
{\em The Effect of Individual Logo Transform.}
We evaluated the impact of different logo image transforms. 
To that end, we eliminated selectively each specific transform in
synthesising the training images and then evaluated any change in model performance.
Table \ref{tab:eval synth data different transformation} shows that:
(1) With synthetic training images alone, 
all geometric and colour transforms except rotation bring some
benefits, especially shearing and scaling. 
The little difference made by rotation makes sense considering that
logo objects appear mostly without any rotation in real scenes.
(2) The benefit of each transform remains after the detector 
is fine-tuned on real training data.

\section{Conclusion}\label{Conclusion}
In this work, we described a new Synthetic Context Logo (SCL) training
image generation algorithm
capable of improving model generalisation capability in deep learning
a logo detector when only sparse manually labelled data is available.
We demonstrated the effectiveness and superiority of the proposed SCL on performing 
logo detection on unconstrained images, e.g., 
boosting relatively the detection accuracy of state-of-the-art Faster
R-CNN network model by $>$$10\%$ on FlickrLogo-32 and $>$$40\%$ on the
more challenging TopLogo-10 benchmark datasets.
Importantly, this performance boost is obtained without the need for
additional manual annotation. It shows the effectiveness of expanding
training data through synthesising pseudo data especially {\em with rich logo context variations}.
As such, 
deep detection model can be learned more reliably with better robustness against
complex background clutters during model deployment.
We carried out detailed evaluation and analysis
on different strategies for model training.
We further introduced a new logo dataset TopLogo-10, 
consisting of top 10 most popular clothing/wearable logos in
challenging visual scene context, designed for more realistic testing
of logo detections in real-world applications.

{\small
\bibliographystyle{ieee}
\bibliography{record}
}

\end{document}